\documentclass[letterpaper]{article} 
\usepackage[pdftex]{graphicx} 
\usepackage{lineno}
\usepackage{amsmath}
\usepackage{hyperref} 

\usepackage[left=0.98in, right=0.98in, top=1.0in, bottom=1.0in]{geometry}

\newcommand\beq{\begin{equation}}
\newcommand\eeq{\end{equation}}

\newcommand\bmat{\begin{bmatrix}}
\newcommand\emat{\end{bmatrix}}

\begin{document}
\setpagewiselinenumbers        
\modulolinenumbers[1]          


\title{Simulation Results on Selector Adaptation in Behavior Trees}

\author{
        Blake Hannaford, Danying Hu, Dianmu Zhang, Yangming Li\\
        Biorobotics Laboratory\\
        Department of Electrical Engineering \\
        The University of Washington
}

\date{\today}

\maketitle

\paragraph{Abstract}
Behavior trees (BTs) emerged from video game development as a graphical language for modeling
intelligent agent behavior.  However as initially implemented, behavior trees are static plans.  
This paper adds to recent literature exploring the ability of BTs to adapt to their success or failure
in achieving tasks.   The ``Selector" node of a BT tries alternative strategies (its children) and returns success only if all of its children return failure.  This paper studies several means by which Selector
nodes can learn from experience, in particular, learn conditional probabilities of success based on 
sensor information, and modify the execution order based on the learned iformation.  
Furthermore, a ``Greedy Selector" is studied which only tries the child having the
highest success probability.   Simulation results indicate significantly increased task performance, 
especially when frequentist probability estimate is conditioned on sensor information.   The 
Greedy selector was ineffective unless it was preceded by a period of training in which all children 
were exercised.

\section{Introduction}
 
Behavior trees (BTs) emerged from video game development as a graphical language for modeling
intelligent agent behavior \cite{halo,lim2010evolving}. BTs have advantages of modularity and scalability with respect to finite state machines.  
When implementing intelligent behavior with BTs, 
the designer of a robotic control system breaks the task down into modules (BT leaves) 
which return either ``success" or ``failure" when called by parent nodes.  All higher level nodes 
define composition rules to combine the leaves including: Sequence, Selector, and Parallel node types. 
A Sequence node defines the order of execution of leaves and returns success if all leaves succeed in order. 
A Selector node (also called ``Priority" node by some authors) tries leaf behaviors in a fixed order, 
returns success  when a node succeeds,  and returns failure if all leaves fail.
Decorator nodes have a single child and can modify behavior of their children with rules such as ``repeat 
until $X>0$".
BTs have been explored in the context of humanoid
robot control
\cite{marzinotto2014towards,colledanchise2014performance,bagnell2012integrated} and as a modeling language for intelligent robotic surgical procedures \cite{hu2015semi}.

\section{Smart BTs}

So far, behavior trees (BTs), as illustrated by the previous examples, have 
been statically created objects, authored by humans to encode an interesting or 
successful high level plan or algorithm.  This paper will explore adaptation 
of BTs to improve their effectiveness at the high level task.   We assume the 
existence of a set of BT leaves, $L_i$, modules of behavior with a 
probabilistic degree of successful completion $P_{Si}$ and, a robust 
means of detecting when each leaf succeeds or fails (which is inherent in all BTs).

\subsection{Literature}
Only a few papers have been published on learning or self-modifying 
BTs\cite{Pereira2015, Colledanchise2015}.   These two papers divide into two approaches, 
improving a pre-defined BT\cite{Pereira2015}, and secondly synthesizing and pruning a BT
from scratch\cite{Colledanchise2015}.  This paper addresses the former challenge, 
improving the performance of an existing BT without changing its topology. 

\subsection{Sequence Nodes}
We first consider the sequence node.  The probability of success for a sequence node
is given by
\beq\label{min_eqn}
P = \Pi_i(P_{Si})
\eeq 
We assume that the leaf composition (selection of leaves) and the order of a 
sequence node cannot be meaningfully changed (i.e. that the selection 
of leaves and their order is required by the task) so that there is little room 
for adaptation of the BT.   However we can still update a local probability estimate
inside the leaves according to 
(\ref{update_eqn}) and use (\ref{min_eqn}) to estimate the overall probability 
of the sequence node succeeding.  This estimate could in turn be used by a sequence node 
higher up in the tree (see below).
Because they are more difficult to adapt, we will not consider adaptive sequence nodes 
further in this paper.  

\subsection{Selector Nodes}

An obvious way that a BT may perform sub optimally is the case of a Selector 
node with several behaviors with different probabilities of success. Since the 
Selector node tries the leaves in left-to-right order (pre-order traversal) a 
lot depends on the order that leaves are placed in the BT.  

Clearly, as in \cite{Pereira2015, Colledanchise2015}, the leaves should be sorted 
in descending order of $P_{Si}$  so that the Selector node tries the most likely to 
succeed node first and the least likely to succeed node last.

It may well be the case that the $P_{Si}$ are unknown at the time the BT leaves 
(action nodes) are created. In this case, we can estimate the $P_{Si}$ on-line 
through experience.   Such an estimate can be created by a frequentist ratio of

\beq\label{update_eqn}
P_{Si} = \frac{\# successes}{T_i}
\eeq
where $T_i$ is the number of trials in which $P_{Si}$ is used.

The probability that a Selector node will succeed is the complement of the
probability that all its leaves fail:

\beq
P = 1 - \Pi_i(1 - P_{Si})
\eeq

\subsubsection{Selector Node Adaptation}

A straightforward optimization of Selector nodes is to update the leaf order as 
new estimates of $P_{Si}$ are generated so that the leaf with the highest 
probability estimate is tried first. Note that if leaves exist with $P_{Si} 
\approx 1$, then subsequent leaves in the pre-order traversal may never be used 
($T_i = 0$).   Such leaves can be optionally pruned from the tree after a 
suitable number of trials.

Note the the updating of $P_{Si}$ can be done strictly locally based on 
(\ref{update_eqn}), or could be done in the cloud based on many robots 
simultaneously working.  During initial learning of the $P_{Si}$, leaf order can 
be randomized to generate non zero $T_i$ for each leaf.

More realistically, certain leaves may be successful in one situation while 
other leaves would be successful in a different situation.   In this case, the 
$P_{Si}$ updating should be conditioned on a feature vector based on sensing 
of the world state.  Suppose that a discrete feature vector, $F(t)$  is available which 
is derived from sensor streams, robot state, and sensor/state history.   This 
feature vector could have its dimension reduced by means of singular value 
decomposition, vector quantization or similar dimensionality reduction 
techniques.

Now we can try to identify clusters in the feature space where each BT leaf 
might be successful or might fail.  Thus we could use a discriminant 
function to predict from the feature space whether or not each leaf will 
succeed in a given situation. Each leaf can be given an estimate of success 
according to the discriminant function.  Alternatively, a success-probability 
distribution, $P(S_i|F_j)$ can be conditioned on the identified clusters ($F_j$) 
of the feature space. Now the smart Selector node could reorder 
the leaves to place  first (leftmost) those with the highest success estimate.  
The $P_{Si}$ is thus effectively conditioned on the feature vector.

\subsubsection{Greedy Selectors}
Although it is intuitive that we can increase performance by reordering the 
Selector leaves, if we are continuously updating leaf probability, it is not clear why
we should tick lower ranked leaves at all.   For example, if the success 
probabilities of leaves L1 - L3 are \{0.8, 0.1, 0.05\}, shouldn't we get better results
by repeatedly ticking L1 even if it fails?

The ``greedy" Selector ticks only the highest ranked child (according any chosen metric) 
and returns its success or failure.   
The greedy Selector can be combined with a 
decorator node to do this until success with a maximum of $N$ tries. 
However, if the metric depends on experience, it is unclear that a Greedy Selector will robustly
discover the leaf metrics.

\subsubsection{Imperfect Success/Failure Judgments}
We now introduce the case where leaves make imperfect judgments of the success 
or failure of their outcome.  Suppose a leaf
attempts a task, and detects success but actually fails. For example, a floor 
cleaning operation could complete a sweep of a room, but leave some dirt on the 
floor.  Assume that another leaf (which could include a human operator) has a 
reliable means for checking success/failure of the previous leaf.   A BT node 
introduced by us, the ``Recovery" node \cite{hu2015semi}, 
returns the system to the proper initial 
state (i.e. moves the robot to the starting corner of the room) and restarts 
the previous leaf.  The true probability of success for the leaf can be 
estimated by the recovery node.

\subsection{Learning Node}

\begin{figure}\centering
    \includegraphics[width=100mm]{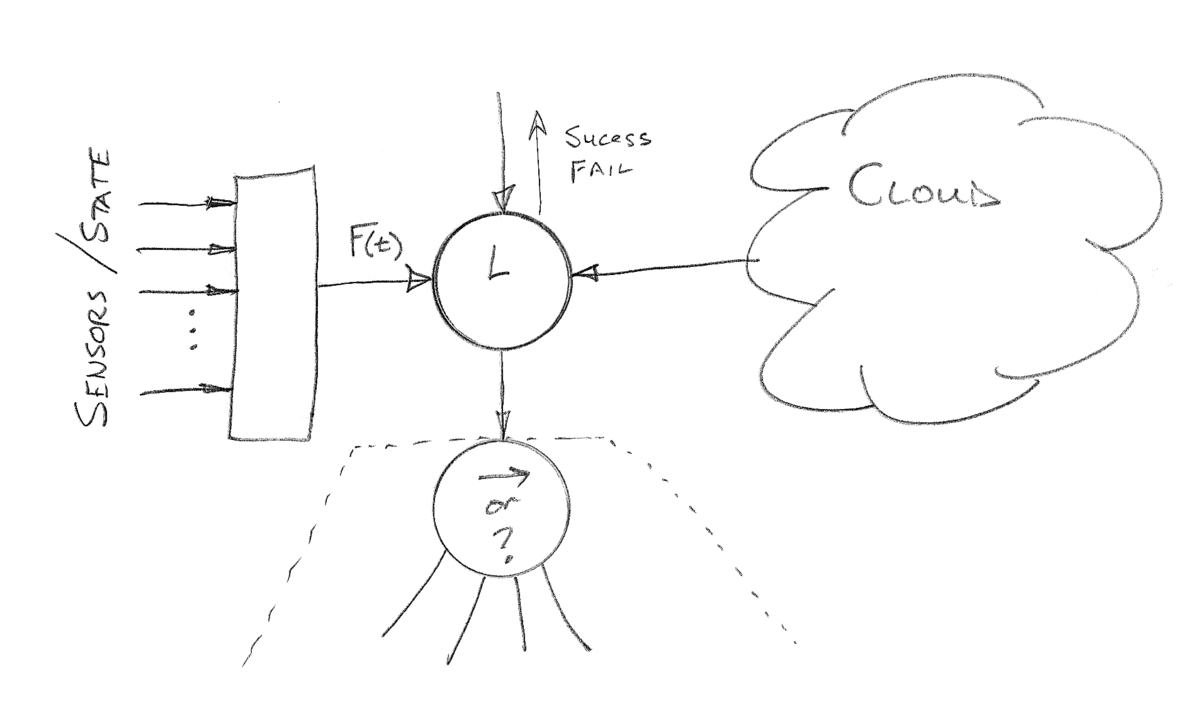}
    \caption{Learning Node which can be attached to any BT node.}
    \label{BehaviorTreeLearningNode}
\end{figure}

As an alternative to adding learning functions to existing node types, 
we can create a new node, the `L' (Learning) Node (Figure
\ref{BehaviorTreeLearningNode}).
The L node works as follows.  Using the history of its single child node (which 
in general is any sub-tree of the BT), the L node accumulates Success and 
Failure outcomes from the child and accumulates an estimate of the probability
\beq
P(S|F)
\eeq
$P(S|F)$ can be estimated in a number of was including Parametric (Gaussian 
or mixture of Gaussians over the feature space), or particle filters.   The L 
node can also obtain data for the estimate from a cloud connected population of 
similar robots having the same child node in their behavior trees and situated 
in similar environments.   
%
%
%
%

\section{Simulations}

\subsection{New Node Features}
A basic simulator was developed using the {\tt behavior3} python BT library\cite{Pereira2015} 
(github.org).  Several new features were added to the node base class:
\begin{itemize}
    \item   Node keeps track of how many times it has been ticked
    \item   Node keeps track of how many times it has reported success. 
    \item   Node has a method to compute $P(S) = N_{success}/N_{ticks}$.  If 
    $N_{ticks}==0$, $P(S) = 0.5$.
    \item   Node keeps an array $N_{t2}$ which stores the number of ticks in the 
    presence of each value of the sensing feature vector $F$.
    \item   Node keeps an array $N_{s2}$ which stores the number of successes in 
    the presence of each value of the sensing feature vector $F$.
    \item   Node has a method to return an array of estimated probabilities:
    \beq
    P_{ij} = \frac  {N_{s2}[j]}  {N_{t2}[j]} = \hat{P}(S_i|F_j)  
    \eeq 
    If $N_{t2}[j]==0$, $P(S_i|F_j) = 0.5$.
    
    \item   Every node has an associated ``Cost," $C$,  for each tick.  Cost for a node can be zero. 
    Cost for each leaf node is determined either by the system designer, or could be calculated, for example
    by an energy meter built into the robot.   A higher level node can sum up the cost of each of its children.
    Although we will not address it in this paper, the Cost estimate can also be conditioned on the sensor cluster, $F$.
    
    \item  Node has a notion of ``Utility" at each tick, $t$.  Utility could be computed in one of 
    two modes: ``Ratio" mode in which Utility is 
    \beq\label{RatioMode}
    U = \frac {P(S)}  {C}  \qquad \mathrm{or} \qquad  \frac {P(S|F)}  {C} 
    \eeq
    and ``Negative Cost" mode: in which Utility is
    \beq\label{NegCostMode}
    U = - P(S) C \qquad \mathrm{or} \qquad  -P(S|F) C
    \eeq
\end{itemize}

\subsection{Adaptive Selector Types}
Using the new node features, we developed several new types of Selector nodes:
\begin{itemize}
        \item {\bf S1. }   reorder the leaves according to $P(S)$. 
        \item {\bf S2. }   reorder the leaves according to $P(S_i|F_j)$
        \item {\bf S2g. }  same as S2, but with Greedy behavior.
        \item {\bf S3. }   reorder the leaves according to $C_i$
        \item {\bf S4. }   reorder the leaves according to $U_i$
        \item {\bf S5. }   same as S4, but compute utility using $P(S_i|F_i)$
    \end{itemize}
    In addition, we define node type {\bf S0} as the regular ``dumb" Selector node.

\subsection{Results}

\begin{figure}\centering
    \includegraphics[width=1.5in]{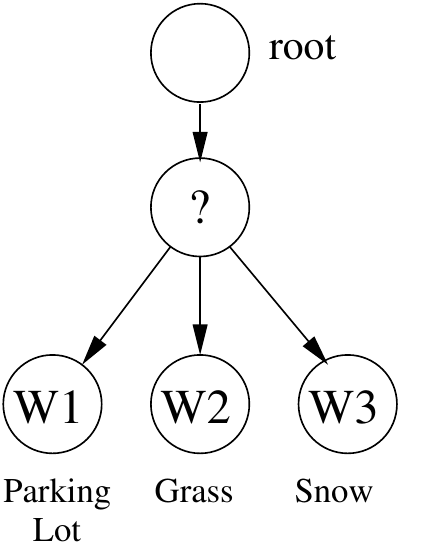}
    \caption{BT for robot walking over simulated terrain. Three different 
        stepping behaviors (1,2,3) have different effectiveness for different
        terrains. They are coordinated by a Selector node (labeled ?).}\label{WalkingTerrainBT}
\end{figure}

A  basic, probabilistic simulator was constructed for the task of a 
humanoid robot walking across several terrains: e.g. Snow, Grass, and Parking Lot.   The BT (Figure 
\ref{WalkingTerrainBT}) selects between three 
``stepping" behaviors (W1 --- W3) and a ``Physics Matrix" gives the probability of a 
successful step by each stepping behavior on each terrain.  It is assumed
that the robot can recover from a failed step but makes no forward progress in 
doing so. 

The test consisted of 250 simulated traversals of a test track consisting 
of 250 steps.   The test track 
consists of stretches of various lengths of each terrain.   The initial Physics 
Matrix was:
\beq\label{physicsmatrixequation}
Ph(n,F) = 
\begin{bmatrix}
    0.8 & 0.1 & 0.1 \\
    0.1 & 0.8 & 0.1 \\
    0.1 & 0.1 & 0.8 \\
\end{bmatrix}
\eeq
The Physics Matrix $Ph(n,F)$ indicates for example that node W1 has a 0.8 
probability of a successful step on terrain 1.  Note that although 
$Ph(n,F) \leq 1$
is a requirement, there is no constraint on the rows or column sums of $Ph$.
For {\bf S4} and {\bf S5}, the cost of each walking method (leaf) 
was set up as follows: $C$(W1) = 4, $C$(W2) = 2, $C$(W3) = 1. 

The measures of success were the inverse number of ticks required to traverse the 
test track, and the total average cost per tick.   The derived metric of overall average Utility was also
computed.  The number of ticks was totaled  and the ticks required at each step 
was averaged over  the 250 trials.  The course was unchanged over the 250 trials. 

\begin{figure}
\includegraphics[width=3.5in]{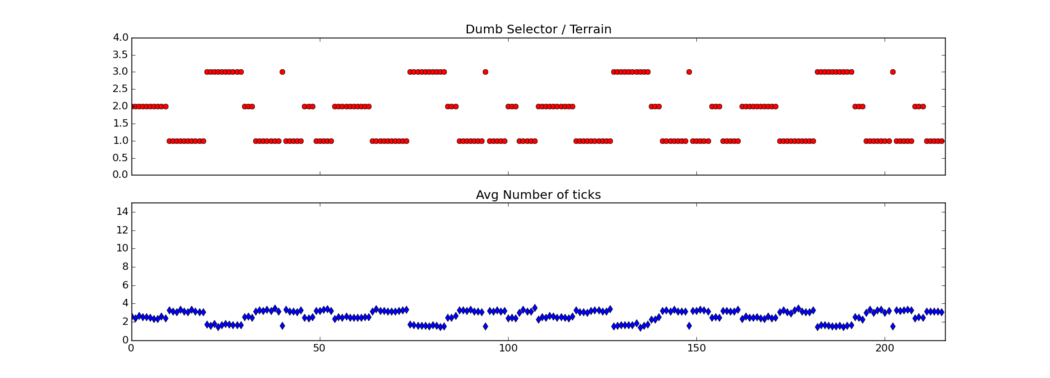}
\includegraphics[width=3.5in]{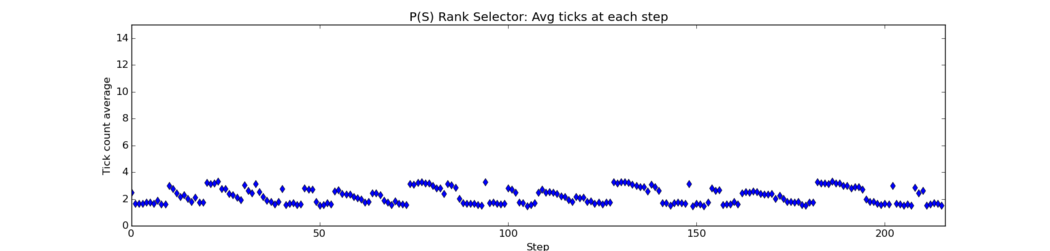}
\includegraphics[width=3.5in]{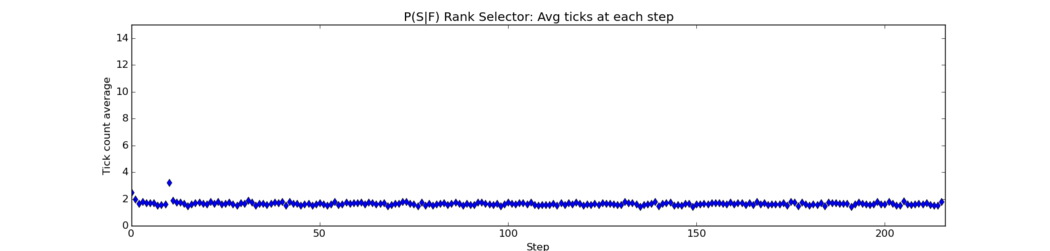}
\includegraphics[width=3.5in]{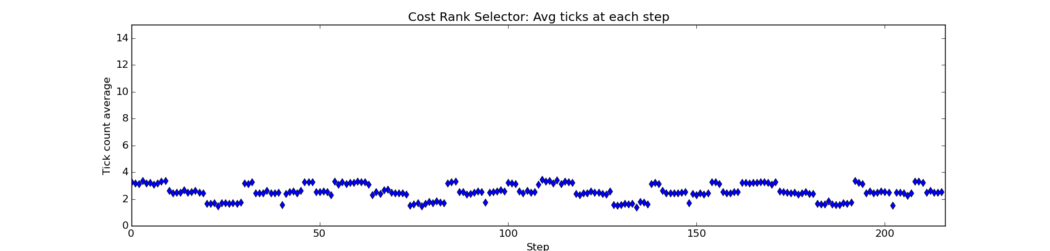}
\includegraphics[width=3.5in]{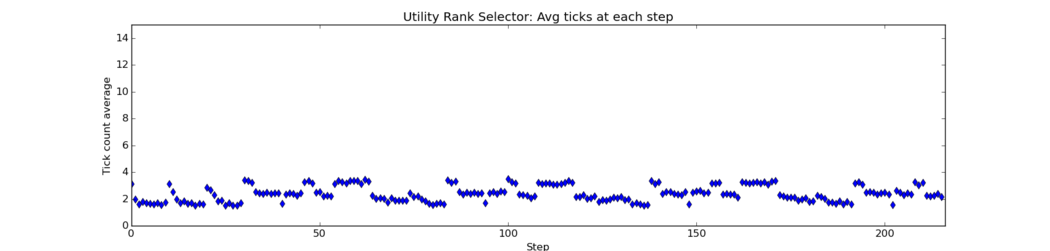}
\includegraphics[width=3.5in]{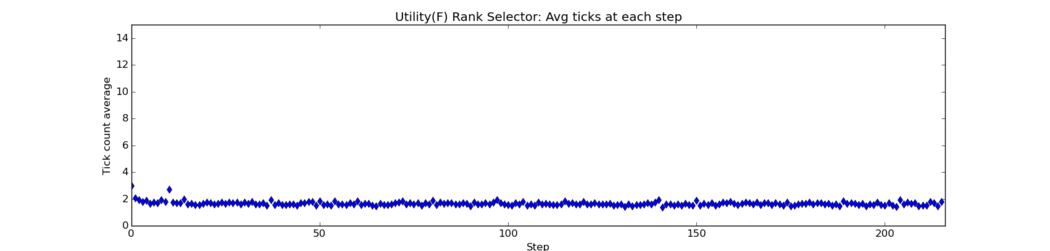}
\caption{Simulation runs on the walking humanoid robot task for each of the five Selector types.
Terrain sequence (red) illustrates the fixed 
sequence of three terrain types. For each simulation result, (blue), X axis is traversal distance in steps.  Y axis is average number of ticks required to complete each step. Terrain sequence is illustrated in uppermost plot.}\label{simplots}
\end{figure}

\begin{table}\centering
\begin{tabular}{l|r|r|r}
    &                 avg ticks     &   avg cost  & avg Util/tick x 1000 \\ \hline
S0: dumb Selector    &    577       &    1286     & 18.25   \\
S1: P(S) order       &    478       &    1151     & 18.51   \\ 
S2: P(S$|$F) order   &    358       &     663     & 36.73   \\ 
S3: Cost order       &   549        &    1031     & 22.62   \\ 
S4: Utility order    &   522        &    1026     & 22.47   \\ 
S5: Utility(F) order &   361        &     844     & 36.57   \\
S2g00: P(S$|$F) Greedy  &  490      &     911     & 59.17   \\
S2g25: P(S$|$F) Greedy  &  284      &     675     & 49.60   \\
\end{tabular}
\caption{Simulated walking performance averaged over the 250 trials for 
each smart Selector (utility values are multiplied by 1000).}\label{simtable}
\end{table}

The results of simulations in which each Selector type traversed the course 250 times are given in Figure \ref{simplots} and Table \ref{simtable}.

\paragraph{S0: ``Dumb" Selector}  The average number of ticks remains the same for each terrain level and does
not adapt over time as expected from its fixed leaf execution sequence.  
With the dumb Selector, the robot took an average of 577 ticks to traverse the course
length of 216 steps, accumulated an average cost of 1286, and an average utility of 
0.01825 on each tick.  These values serve as a baseline for comparison with the
smart Selectors (below).

\paragraph{S1: P(S) Selector} The first smart Selector reorders its leaves in descending order of their overall
probability of success.  It can be seen that the number of clicks adapts over the time that terrain remains constant but then increases at each terrain change. 

The average number of ticks per course traversal was substantially reduced to 478.
Cost and Utility were approximately the same.

\paragraph{S2: P(S$|$F) Selector}  This Selector tries its leaves in descending order of their probability of success conditioned on a simulated sensor measurement of the terrain type (which is 100\% accurate). The average tick count quickly adapts to a steady state value of a little less than two clicks per step. Note transients of worse performance are observed at the first transitions between terrain types. 

Average ticks per traversal for S2 were the lowest of all at 358.  Cost and 
utility were also the best. 

\paragraph{S3: Cost Selector} This Selector executes its leaves in increasing order of leaf cost.   This should not appreciably affect click rate compared to the dumb Selector, but different average click rates are associated with each
terrain type because the cost ranking is different than the initial order. 

The Cost-based Selector had significantly lower cost (1031) than the dumb Selector,
but the performance of the cost Selector depends on the relative cost assigned to the 
nodes which was arbitrary in this experiment. 

\paragraph{S4: Utility Selector}  Both Utility Selectors (S4 and S5) were initially studied with Ratio mode
(\ref{RatioMode}).  
The Utility Selector combines the fixed cost weights with the experience-based 
probability estimates.  As with S1, the tick count per step adapts downwards during the 
time terrain remains constant. 

Because of the symmetry of the chosen physics matrix, and the roughly equal distribution
of terrain states in the trajectory, $P(S)$ is about the same for 
each leaf and thus the numerical results for utility order are very similar to those of cost order. 

\paragraph{S5: Utility(F) Selector} This Selector derives utility based on the sensor-conditioned success probability.
Similarly to the S2 Selector, it rapidly converges to a steady state value. 

Although S5 has a very low average tick count (361), its overall performance
is not significantly better than S2, even for average Utility.

\begin{figure}\centering
    \includegraphics[width=3.5in]{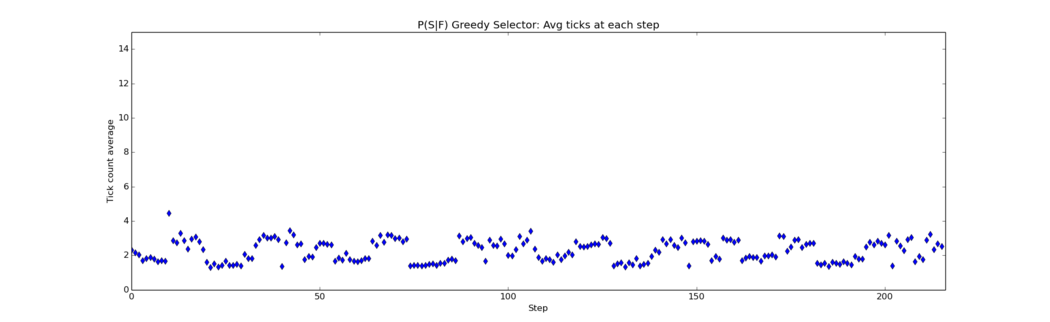} \\
    \includegraphics[width=3.5in]{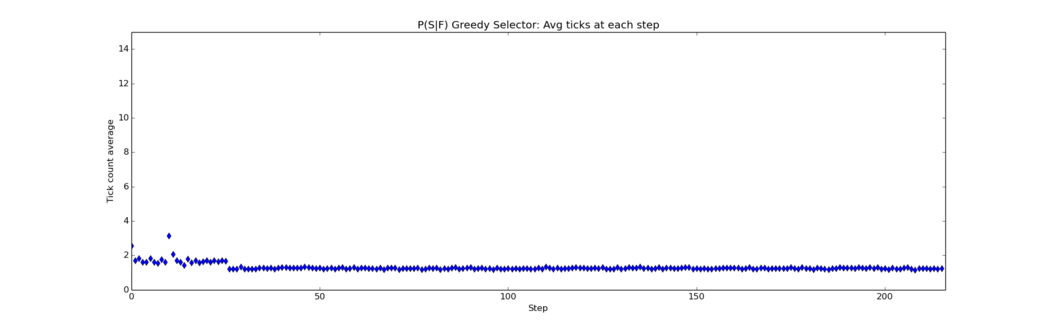}
    \caption{Simulated walking runs with the Greedy P(S|F) Selector.  Performance is 
    poor (top) because all the leaves are not exercised enough to generate 
    robust estimates of $P(S|F)$.  When Greedy mode is enabled after 25 initial 
    steps with the normal S2 Selector (bottom trace) performance is excellent.}\label{greedySelectorplots}
\end{figure}

\paragraph{S2g: Greedy P(S$|$F) Selector)}
This Selector repeatedly ticks only the highest probability leaf.   The performance of this 
Selector is not good (Figure \ref{greedySelectorplots}, top).  Because it is constantly selecting 
only the highest probability, it is not exercising the other leaves enough for them to have robust
probability estimates.  Another set of 250 runs was performed in which the Greedy mode was not enabled until 25 steps were completed with the normal S2 Selector.   This mode allowed
the probability estimates to converge.  Subsequently the greedy mode performed extremely well.

\subsubsection{Ratio vs Linear Utilities}
All Selectors were re-run with the Utility computation changed to Negative Cost mode (\ref{NegCostMode}). 
The average tick counts and average cost values were essentially unchanged (max difference 5 ticks/ 9 cost points).
The utility values were not directly comparable, but the ranking of highest to lowest utility changed 
dramatically, in Ratio Mode, the highest two Selectors for Utility were S2g and S4 (Greedy P(S|F) and 
Utility(F) order).   In Negative Cost Mode, the highest two Selectors were S3 and S4 (Cost order and 
Utility order) and number 3 was S5 (Utility(F) order).   
The relative utility achieved by the Selectors is thus highly sensitive to the Utility definition. 

\subsubsection{Cost Weights}
We investigated the sensitivity of these results to choice of costs on the three nodes which were initially [2,4,1] for the three nodes respectively.  
We re-ran the tests above with the costs 
reset to [2,1,4].   The average tick counts were mostly unchanged.  The largest difference was 
5 ticks  (out of 515) except for S3, Cost order, which decreased by 32 ticks on average when the costs were changed.  

Average cost per run was less consistent.   The average cost increased for S0 (dumb Selector) and 
for S2g00 (Greedy Selector without training interval), but decreased substantially for the others. 
Average Utility decreased for the same two Selectors, but increased for the others. The relative
performance of the various Selectors remained approximately the same except for a drop in ranking
of the S2g00 (untrained Greedy) Selector.

\subsubsection{Symmetry of Physics Matrix}

In the previous tests, the Physics matrix was given by (\ref{physicsmatrixequation}).  Using the 
first set of cost weights:  [2,4,1] , we changed the physics matrix to

\beq\label{newphysicsmatrixequation}
Ph(n,F) = 
\begin{bmatrix}
    0.6 & 0.1 & 0.1 \\
    0.4 & 0.7 & 0.4 \\
    0.1 & 0.1 & 0.8 \\
\end{bmatrix}
\eeq
to introduce better and weaker nodes.   Node two is now reasonably good at all terrains although not the best on terrains 1 and 3. 

When this change was introduced to the system, average tick counts per run went up for all 
Selectors except the dumb Selector.  Cost also increased for all Selectors.  Average
utility declined modestly for all Selectors.

\subsubsection{More Complex Scenario}
\begin{figure}\centering
    \includegraphics[width=3.5in]{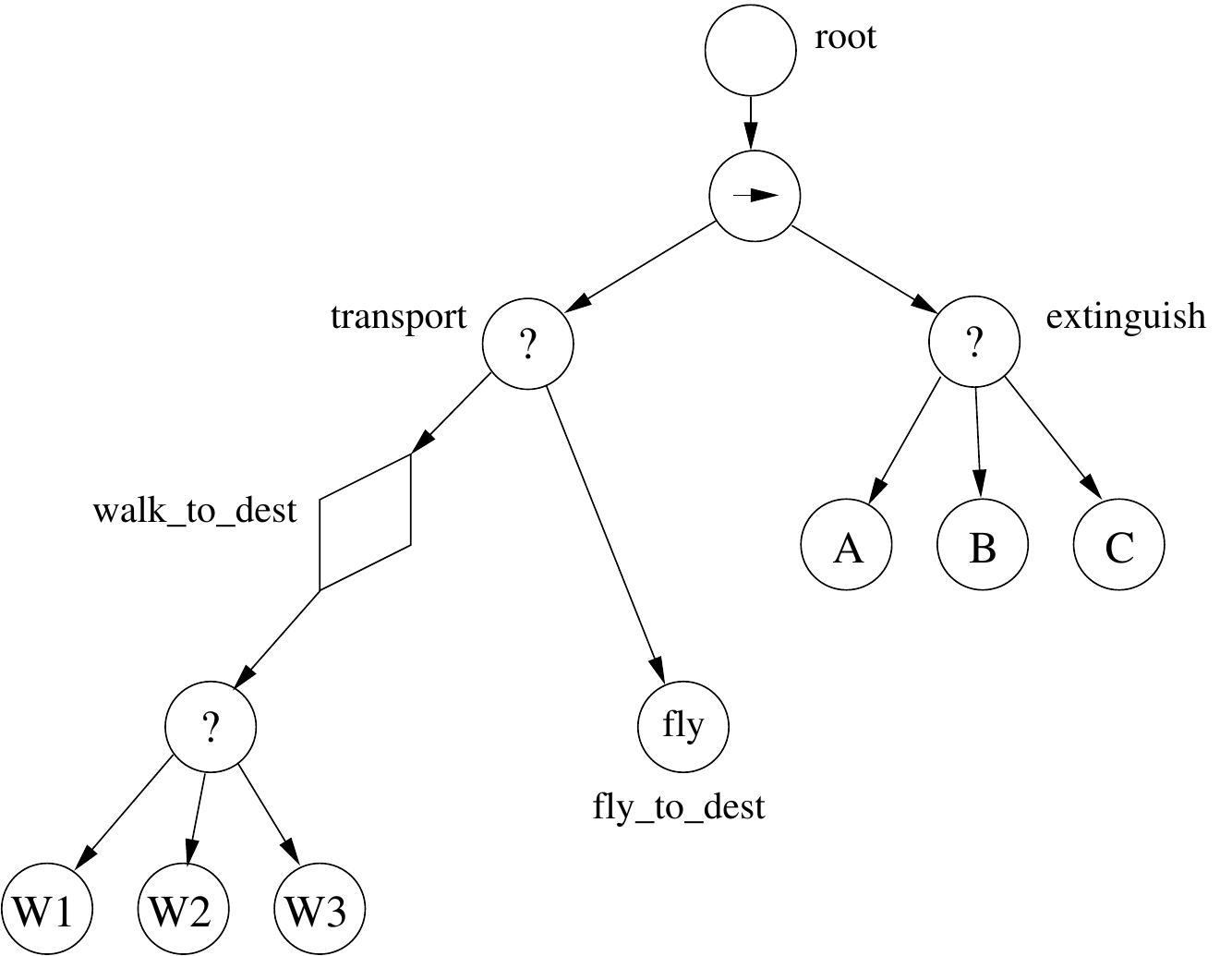} 
    \caption{Behavior tree for fire rescue task (adapted from \cite{Pereira2015}).}\label{firerescueBT}
\end{figure}

A more complex scenario derived from the fire rescue scenario 
studied in Pereira and Engel\cite{Pereira2015} was simulated.  
In our modified version of this scenario (Figure \ref{firerescueBT}), 
the robot walks to a fire (in the same was as the walking scenario above).  
The distance to the fire is 100 steps.  If the robot fails to reach the fire in {\tt walk\_limit} steps, 
the robot flies to the fire using its jet-pack (at much higher cost).  
The cost of a step is set to one and the cost of flying to the destination is 350.   Flying succeeds with 
probability 0.9.   A value of the {\tt walk limit} at which the robot must convert to flight about 50\% of the time 
with the dumb Selector was experimentally determined to be 120 steps.  Simulations were also run with higher
{\tt walk\_limit}s to reduce the number of flight conversions.   

At the fire, there are three fire types.  The robot uses one
of three extinguishers with  a fire fighting physics matrix of 

%
%

\beq\label{firephysicsmatrixequation}
Pf(n,F) = 
\begin{bmatrix}
    1.00  & 0.05 & 0.05 \\
    0.05 & 1.00 & 0.05\\
    0.05 & 0.05 & 1.00 \\
\end{bmatrix}
\eeq

The interpretation of the fire physics matrix is that extinguisher A has a 100\% probability of extinguishing
fire type A, and a small probability of extinguishing fire types B and C, etc.   The intent of setting 
at least one element in each column to 1.0 is so that success of the tree as a whole depends only on  the transport node.

\begin{table}\centering
\begin{tabular}{l|r|r|r|r|r|r|r}
    &          limit &        avg ticks     &   avg cost  & avg Util  & walk ticks &  fly ticks  & fails \\ \hline
Sel00: Dumb Selector &150    &    228       &    245   &   0.41    &   1000     &       0    &     0 \\
Sel00: Dumb Selector &120     &   224       &    441   &   0.28    &   1000     &      574   &    52 \\ \hline
Sel01: P(S) ranking &120       &   22       &    388   &   0.26    &     89     &     1000   &    45 \\ \hline
Sel02:  P(S$|$F) ranking &120    &   18       &    376   &   0.27    &    103     &     1000   &    63 \\
Sel02:  P(S$|$F) ranking &150    &   18       &    376   &   0.27    &     97     &     1000   &     0 \\
Sel02:  P(S$|$F) ranking &900   &   167       &    176   &   0.57    &   1000     &       1    &     0 \\\hline
Sel03: Cost ranking &120      &  165        &    379   &   0.35    &   1000     &      568   &    71 \\\hline
Sel04: Utility ranking &120   &     28      &    393   &   0.26    &    117     &      999   &    60 \\\hline
Sel05: Utility(F) ranking &120  &   16      &    373   &   0.27    &     89     &      999   &    38 
\end{tabular}
\caption{Simulated fire fighting scenario. ``limit" is the number of ticks allowed to reach fire before
    robot switches to flight. 1000 trials per Selector.   }\label{firesimtable}
\end{table}

\paragraph{Results}
The firefighting task of Figure \ref{firerescueBT} was simulated 1000 times for each BT and the {\tt walk\_limit}
was changed to understand the frequency of conversion to flying.   

\paragraph{S0} S0 had no conversions to flight when {\tt walk\_limit} was 150, but about 57\% flight conversions
with {\tt walk\_limit} at 120. The average cost increased by 196 ticks as the number of expensive flights increased
(flight cost = 350).  Failures increased to about 10\% of flights since the flight mode had only 90\% success
probability.  All walking trials were overall successful since extinguish node is guaranteed to succeed. 

\paragraph{S1}  S1  (P(S) ranking) was evaluated with {\tt walk\_limit} of 120 in which the probability of successfully walking 
to the fire is about 50\%.  In this case the $P(S)$ Selector adapted to the higher success rate of flying 
and flew on all 1000 trials.  Average cost was still lower than the equivalent dumb Selector trials, presumably
because of fewer wasted steps prior to flying. 

\paragraph{S2}  S2 strongly preferred flight unless the {\tt walk\_limit} was set to the very high value of 900. 

\paragraph{S3}  S3 performed similar to the Dumb Selector although with slightly lower cost.

\paragraph{S4 and S5}  Utility ranking was evaluated in Ratio mode.   Both Utility ranking modes performed similarly to the S2 ($P(S|F)$) ranking.

\subsection{Simulation: Conclusions}

\subsubsection{Terrain Navigation}
Smart Selectors performed robust adaptation to the simulated walking task under a variety of
circumstances.  From the point of view of how many ticks were required to navigate the terrain,
the worst Selector under all circumstances was the dumb Selector (and its close
relative the untrained Greedy Selector).  
The two best Selectors were the S2g25 (trained greedy) Selector, and the S2 $P(S|F)$ Selectors under all conditions tested.  In Ratio Mode, the two Selectors
designed to minimize cost or maximize utility did not do so relative to the two 
best performing nodes. When measured by average tick count per traversal, 
the performance of the Cost Selector was most sensitive to change in the node costs as expected. 

However, when the Utility measure was changed to Negative Cost Mode (\ref{NegCostMode}), S3-S5, the 
Selectors which chose Cost and Utility had the highest average Utility, and the Dumb Selector and the 
untrained greedy Selector the worst.  This suggests a linear Utility measure operates more intuitively. 

\subsubsection{Firefighting}
The more complex fire-fighting scenario was simulated in order to study the extension of the
Selectors tested in the terrain scenario to a slightly more complex task.  It is less straightforward to 
predict the effects of smart Selectors as the scenario gets more complex, but increasing the {\tt walk\_limit}
simulation parameter, which increases the chance of successfully walking to the fire, produced the 
expected changes in tick counts, and total costs. 
 
\section{Discussion}
This paper has explored adaptation in Behavior Tree Selector nodes through some simulation experiments.   
The simulations showed  that adaptive selectors could considerably improve the performance of Selector nodes
through a variety of methods.  Evaluating node probabilities conditioned on quantized sensor information was 
effective.  Measures of cost and utility were less  effectively optimized by experience through behavior trees. 

A greedy selector, which selects only its node of highest probability, was potentially very effective, but 
only when activated after a period of learning in which all leaves were activated enough times to develop
robust probability estimates. 

We plan to further explore these results with physical experimental platforms, and to explore their applicability
to medical robotics. 

\newpage
\paragraph{Acknowledgements}
    We are pleased to acknowledge support for this work from
the Korean Institute of Science and Technology (KIST, Dr. Hujoon project), 
the U.S National Institutes
of Health NIBIB R01 EB016457 NRI-Small,
and the National
Science Foundation (grant number 1227406) under a subcontract
from Stanford University.

\bibliography{adapt_BTs_icra}
\end{document}